\documentclass[sn-mathphys-num]{sn-jnl}% Math and Physical Sciences Numbered Reference Style 

\usepackage{graphicx}%
\usepackage{multirow}%
\usepackage{amsmath,amssymb,amsfonts}%
\usepackage{amsthm}%
\usepackage{mathrsfs}%
\usepackage[title]{appendix}%
\usepackage{xcolor}%
\usepackage{textcomp}%
\usepackage{manyfoot}%
\usepackage{booktabs}%
\usepackage{algorithm}%
\usepackage{algorithmicx}%
\usepackage{algpseudocode}%
\usepackage{listings}%

\usepackage[most]{tcolorbox}

\begin{document}
	
	\title{Automating Pharmacovigilance Evidence Generation: Using Large Language Models to Produce Context-Aware SQL}
	
	%%=============================================================%%
	%% Authors
	%%=============================================================%%
	
	\author*[1]{\fnm{Jeffery L.} \sur{Painter}}\email{jeffery.l.painter@gsk.com}
	
	\author[1,2]{\fnm{Venkateswara} \sur{Rao Chalamalasetti}}\email{venkateswararao.x.chalamalasetti@gsk.com}
	
	\author[3]{\fnm{Raymond} \sur{Kassekert}}\email{raymond.x.kassekert@gsk.com}
	
	\author[4,5]{\fnm{Andrew} \sur{Bate}}\email{andrew.x.bate@gsk.com}
	
	\affil*[1]{\orgname{GlaxoSmithKline}, \orgaddress{\city{Durham}, \state{NC}, \country{USA}}}

	\affil[2]{\orgname{Tech Mahindra}, \orgaddress{\city{Plano}, \state{TX}, \country{USA}}}
		
	\affil[3]{\orgname{GlaxoSmithKline}, \orgaddress{\city{Upper Providence}, \state{PA}, \country{USA}}}
	
	\affil[4]{\orgname{GlaxoSmithKline}, \orgaddress{\city{Brentford}, \state{Middelsex}, \country{UK}}}
	
	\affil[5]{\orgname{London School of Hygiene and Tropical Medicine}, \orgaddress{\city{London}, \country{UK}}}
	
	\abstract{
		
		\textbf{Objective:} To enhance the efficiency and accuracy of information retrieval from pharmacovigilance (PV) databases by employing Large Language Models (LLMs) to convert natural language queries (NLQs) into Structured Query Language (SQL) queries, leveraging a business context document.

		\textbf{Materials and Methods:} We utilized OpenAI’s GPT-4 model within a retrieval-augmented generation (RAG) framework, enriched with a business context document, to transform NLQs into syntactically precise SQL queries. Each NLQ was presented to the LLM randomly and independently to prevent memorization. The study was conducted in three phases, varying query complexity, and assessing the LLM's performance both with and without the business context document.

		\textbf{Results:} Our approach significantly improved NLQ-to-SQL accuracy, increasing from 8.3\% with the database schema alone to 78.3\% with the business context document. This enhancement was consistent across low, medium, and high complexity queries, indicating the critical role of contextual knowledge in query generation.
		
		\textbf{Discussion:} The integration of a business context document markedly improved the LLM's ability to generate accurate and contextually relevant SQL queries.  Performance achieved a maximum of 85\% when high complexity queries are excluded, suggesting promise for routine deployment.
		
		\textbf{Conclusion:} This study presents a novel approach to employing LLMs for safety data retrieval and analysis, demonstrating significant advancements in query generation accuracy. The methodology offers a framework applicable to various data-intensive domains, enhancing the accessibility and efficiency of information retrieval for non-technical users.}

	
	\maketitle	

		\section{Introduction}
			
			Drug safety, or pharmacovigilance (PV), involves the systematic assessment of medications and vaccines to ensure that benefits outweigh risks. Central to PV are extensive databases compiling individual case safety reports (ICSRs), which are crucial for risk identification, strategy development, and regulatory reporting \cite{beninger2018pharmacovigilance}.
			
			Navigating safety databases to generate precise queries is inherently complex, requiring specialized expertise due to their vast, multi-table, and interlinked structures. Unlike many other relational databases, safety databases must manage diverse case types, regulatory requirements, and variations in product and event reports, all while ensuring compliance with stringent reporting standards. As noted by Brass and Goldberg, this complexity often arises from the mismatch between ontologies and relational databases \cite{brass2006semantic}, leaving users struggling to articulate accurate search criteria. The prevalence of query misapplication in healthcare further underscores the need for enhanced methods \cite{sivarajkumar2024clinical}. Recent studies demonstrate the effectiveness of combining heuristic reasoning with deep learning for predicting semantic group assignments, achieving high accuracy and potentially supplementing automated query generation tasks \cite{mao2023two}. For example, the CHESS framework introduces a multi-component pipeline leveraging large language model (LLM)-based methods for entity and context retrieval, schema selection, and SQL generation, improving data retrieval in complex real-world databases \cite{talaei2024chess}.
			
			Our research presents a novel approach to converting natural language queries (NLQs) into SQL code for retrieving information from large, complex safety datasets. Safety databases present unique challenges, storing highly diverse data—ranging from case reports to product details, adverse events, and regulatory submissions—while complying with stringent reporting requirements across regions. Our company's safety database contains over 500 tables, with more than 50 columns per table, spanning 50 years of data and encompassing at least 5,000 distinct fields. The largest table holds nearly 1.3 billion rows, and a single safety case may include as many as 100 products and 100 events, adding to the complexity of tracking and querying. To address these challenges, we developed a \textit{business context document} that distills intricate business rules and database knowledge from PV experts into accessible, plain language. Integrating this context document with the LLM enhances the model's ability to craft queries that closely align with business needs, overcoming the limitations of relying solely on database metadata.
			
			While research on NLQ-to-SQL tasks is ongoing, our approach significantly differs from frameworks like CHESS by introducing the business context document. This document enhances accuracy by providing the LLM with domain-specific knowledge, aligning query generation with business rules and database structures. Contextual knowledge is crucial for improving domain-specific NLQ tasks, with applicability beyond PV. Our work addresses the challenge of navigating an extremely large, complex database, a level of complexity not explicitly tackled by the CHESS framework.
			
			Crafting queries in PV datasets requires both technical and scientific expertise, and translating safety scientists' nuanced requests into precise code poses a significant challenge with considerable room for error \cite{taipalus2018errors}. While tools like Query-by-Example (QBE) offer simplified means of crafting database queries \cite{zloof1977query}, they fall short in handling complex needs \cite{ramakrishnan2002database}.
			
			The rise of LLMs in natural language processing has improved data management by enabling more efficient and accurate query generation \cite{dauphin2017language, li2023can, gao2023text}. However, LLMs alone often provide only reasonable, but not perfect, performance in domain-specific tasks \cite{painterLLM2023}. In heavily regulated environments like PV, accuracy is paramount, requiring approaches that integrate the expertise of domain specialists.
			
			Recent evaluations highlight performance gaps between proprietary models like GPT-4 and open-source alternatives \cite{li2023can, roberson2024analyzing, zhang2024structure}. While general query generation tasks have achieved up to 72\% accuracy through advanced retrieval and schema selection techniques \cite{talaei2024chess}, LLM-generated SQL queries can suffer from issues like “hallucinations” and other errors \cite{qu2024before}. This underscores the need for supplementing LLMs with detailed contextual information to improve query precision in PV and other highly regulated domains.
			
			Relying on technical teams for safety query formulation can introduce delays, underscoring the need for solutions that empower safety scientists with direct access to data through LLM-driven tools. Our approach builds on advancements in LLM technology, offering an intuitive interface that allows non-technical users to perform complex data queries, potentially enhancing PV data analysis and reporting \cite{painterLLM2023}. By integrating LLMs with a business context document, we bridge the gap between technical complexity and domain expertise, democratizing access to critical drug safety informatics.

		\section{Methods}
			
			Our study aimed to transform natural language queries (NLQs) into syntactically precise Oracle\textsuperscript{TM} SQL queries, utilizing Large Language Models (LLMs) within a retrieval-augmented generation (RAG) framework, enhanced by a detailed business context document. The experiment was conducted in three phases to assess the LLM's capability in SQL query generation under varying levels of contextual knowledge.
			
			\subsection{Experimental Phases}
			
			The experiment was structured into three phases, each targeting different aspects of LLM performance in translating NLQs into SQL. \textbf{Phase 1} established a baseline using an exhaustive database schema. \textbf{Phase 2} introduced a business context document, which provided plain language descriptions of data structures. \textbf{Phase 3} narrowed the focus to essential tables, aiming to determine whether a more targeted approach could enhance or match the performance observed in Phase 2.
			
\begin{figure}
				\centering
				\includegraphics[width=5.0in]{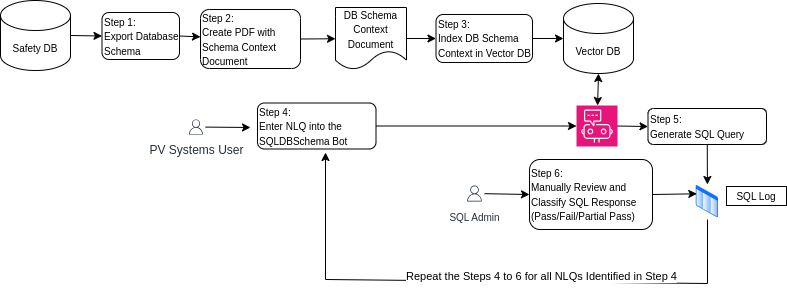}
				\caption{Phase 1 Experimental Design}\label{fig_01}
			\end{figure}

			In \textbf{Phase 1}, as shown in Figure \ref{fig_01}, the LLM was given comprehensive schema documentation, including a 290-page PDF detailing every table and column definition.
			
\begin{figure}
				\centering
				\includegraphics[width=5.5in]{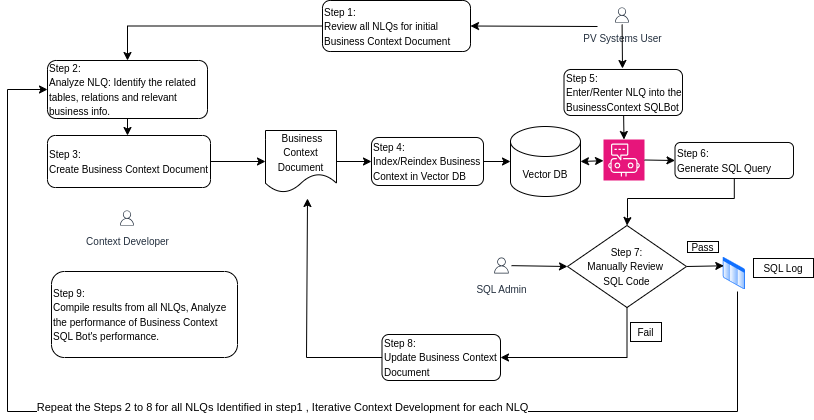}
				\caption{Phase 2 Experimental Design}\label{fig_02}
			\end{figure}

			\textbf{Phase 2} mirrored Phase 1 but introduced a business context document, created by safety data experts. This document summarized key data structures in plain language, providing contextual insights into the relevance of the database elements (Figure \ref{fig_02}).
			
			In \textbf{Phase 3}, the original 290-page document was condensed to 29 pages, containing only 36 essential table definitions. This streamlined approach focused the LLM on the most relevant data, aiming to improve query generation effectiveness.
			
			\subsection{NLQ Selection}
			
			Sixty NLQs were selected from historical user logs, covering a broad spectrum of query complexity. These queries were consistent across all phases and ranged from simple to complex data retrieval tasks.
			
			A structured prompt guided the LLM to generate syntactically correct SQL queries while minimizing unfounded responses. The chatbot prompt defined a specific persona, instructing the model:
			
				\textit{``You are an Oracle SQL expert. Given a question, generate a syntactically correct Oracle SQL query. Avoid querying non-existent columns and pay close attention to column-table associations. For keywords in the WHERE clause, ensure case-insensitive data comparison, for example, `upper(STATE\_NAME) = upper('deleted')`. If you are unable to generate the SQL query, please state that you cannot create the query without additional information or context, do not attempt to make anything up.''}			

			In all phases, a vector-based retrieval strategy, utilizing embeddings from the text-embedding-ada-002 model\footnote{\url{https://platform.openai.com/docs/guides/embeddings/embedding-models}}, was employed. Each phase involved background knowledge tailored to the experiment -- Phase 1 used the full schema, Phase 2 added contextual business knowledge, and Phase 3 focused on essential data elements.
			
\begin{figure}
				\centering
				\fbox{\includegraphics[height=4.0in]{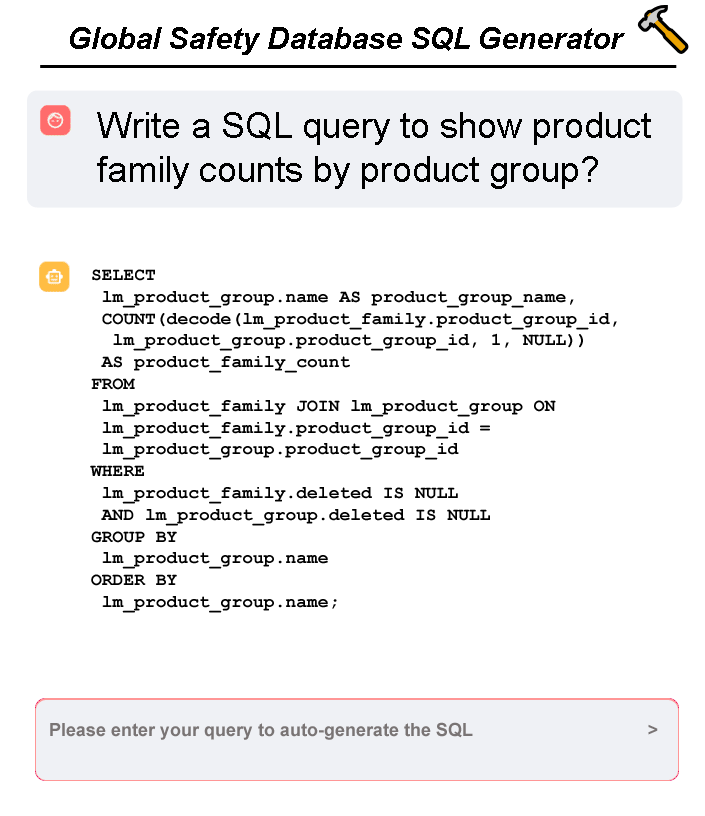}}
				\caption{LLM Chatbot User Interface}\label{fig_03}
			\end{figure}

			Each experiment employed a text splitter (chunk size: 1,000 characters; overlap: 200 characters) to process inputs for the GPT-4 model, as shown in Figure \ref{fig_03}.
			
			\subsection{Complexity Scoring Algorithm}
			
			We developed a scoring algorithm to objectively measure SQL query complexity, which considers factors such as the number of tables, joins, and clauses in the query, as well as the estimated time required for manual creation. This method offers a reproducible way to evaluate SQL complexity, acknowledging that complexity assessments can vary widely in methodology \cite{simon2010sql}, as shown in Algorithm \ref{alg:complexity}.
			
			\begin{algorithm}
				\caption{Compute the SQL Complexity Score}
				\label{alg:complexity}
				\begin{algorithmic}[1]
					\State $\text{score} \gets 0$
					\State $\text{score} \gets time\_to\_create$
					\State $\text{score} \gets \text{score} + number\_of\_tables$
					\State $\text{score} \gets \text{score} + number\_of\_joins$ 
					\State $\text{score} \gets \text{score} + number\_of\_where\_clauses$ 				
					\If{$\text{has\_group\_by} = \text{True}$}
					\State $\text{score} \gets \text{score} + 1$
					\EndIf
					\If{$\text{has\_order} = \text{True}$}
					\State $\text{score} \gets \text{score} + 1$
					\EndIf
					\If{$\text{has\_aggregation} = \text{True}$}
					\State $\text{score} \gets \text{score} + 1$
					\EndIf
					\State \Return $score$				
				\end{algorithmic}
			\end{algorithm}
			
			\subsection{Experimental Setup and Classification of Generated SQL}
			
			Each NLQ was presented to the LLM randomly and independently to prevent memorization. The context was reset between evaluations, ensuring no data carried over between phases. A safety data expert evaluated each SQL query, categorizing them into \textit{pass}, \textit{fail}, or \textit{partial pass}. A ``pass'' indicated the SQL was executable without modifications, a ``fail'' occurred when the LLM couldn’t generate a valid query, and a ``partial pass'' indicated minor modifications were needed.
			
			This classification approach informed the development of our phases, and Phases 1 and 2 evaluated performance with and without contextual knowledge. Phase 3 focused the LLM on essential tables to assess whether this refinement improved performance.
			
			\subsection{Constructing the Business Context Document}
			
			The business context document was developed in Phase 2 and captures domain-specific knowledge in unstructured text form. It prioritized frequently used tables, aligned with regulatory and reporting requirements, and incorporated input from domain experts. This iterative process ensured accuracy, relevance, and completeness. The final document contained 36 key table definitions, along with business rules and relationships necessary for query generation.
			
			Examples of domain-specific guidance included:
			
			\textbf{Definition} \textit{“New case”: A new case includes cases not yet assigned to any database user (e.g., Intake Specialist or Data Entry Specialist) as well as cases assigned but not yet accepted by the user.}
			
			Additionally, explicit SQL generation suggestions were provided, such as:
			\begin{itemize}
				\item \textit{Avoid using SELECT * due to the vastness of some tables.}
				\item \textit{Always join tables using their primary and foreign keys.}
				\item \textit{Consider database performance.}
			\end{itemize}
			
			The business context document was refined iteratively, with minimal ongoing updates, and significantly reduced the time spent on query generation. Updates to database schemas could impact the document's utility, and reliability should be monitored over time.
			
			An example of the contents of our business context document is highlighted below.
			
			\begin{tcolorbox}[colback=blue!5!white,colframe=blue!75!black]
				
				\begin{center}
					\textbf{\textit{Excerpt of the Safety Database Context Document}}
				\end{center}
				
				\textbf{Purpose:} The purpose of this context document is to equip the Large Language Model (LLM) with the
				knowledge and intricacies of the global Safety Database. This will enable accurate and efficient SQL
				generation for data retrieval and analysis. \newline
				
				\textbf{Database Overview:} The global Safety Database is a comprehensive pharmacovigilance software solution
				that stores and manages data related to drug safety and adverse event reporting. \newline
				
				\textbf{Introduction:}
				
				This document provides guidance on generating SQL related to business configurations in the Global
				Safety database system. It emphasizes the tables, relationships, and common querying patterns associated with
				these configurations. \newline
				
				\textbf{Data Tables and Relationships:}
				
				The global safety database contains the following tables:
				
				\begin{enumerate}
					\item 
					
					Table name PRODUCT\_FAMILY contains company product data,
					where the column identified as FAMILY\_ID is the primary key,
					the column identified as NAME contains the product family name,
					and the column identified as DELETED contains the date a record was deleted. \newline
					
					If the DELETED column is NULL, then the record is not deleted. \newline
					
					The column PRODUCT\_GROUP\_ID is a foreign key 
					to link records matching on the same column from
					the table PRODUCT\_GROUP.
					
				\end{enumerate}
				
			\end{tcolorbox}

			An illustrative example of the Business Context document’s utility is when users request follow-up letter counts. The database does not explicitly track letters, but the document specifies that these are stored in case attachment files, using the field \texttt{CLASSIFICATION}. For follow-up letters, the relevant SQL query rule is defined as \texttt{CLASSIFICATION like '\%FU Attempt\%'}, bridging the gap between user queries and the database structure.
			
			\subsection{System Design and Implementation}
			
			LLMs demonstrate proficiency in answering general queries, but they often fall short when required to extract information from specific contexts without domain-specific training \cite{brown2020language} \cite{xie2021explanation}. Fine-tuning LLMs with contextual knowledge, such as business documents, significantly enhances their ability to generate accurate, contextually informed responses \cite{radford2019language}. 
			
			To leverage this capability, we implemented a RAG-based pipeline, integrating vector-based data representations to improve SQL query generation accuracy \cite{lewis2020retrieval} \cite{topsakal2023creating}. Our system uses OpenAI’s GPT-4, hosted on Azure\footnote{\url{https://azure.microsoft.com/}} and integrated with LangChain\footnote{\url{https://www.langchain.com/}}. The system ensures generated SQL queries are accurate, contextually relevant, and accessible to non-technical users.
			
			\begin{figure}
				\centering
				\includegraphics[width=\textwidth, width=6.0in]{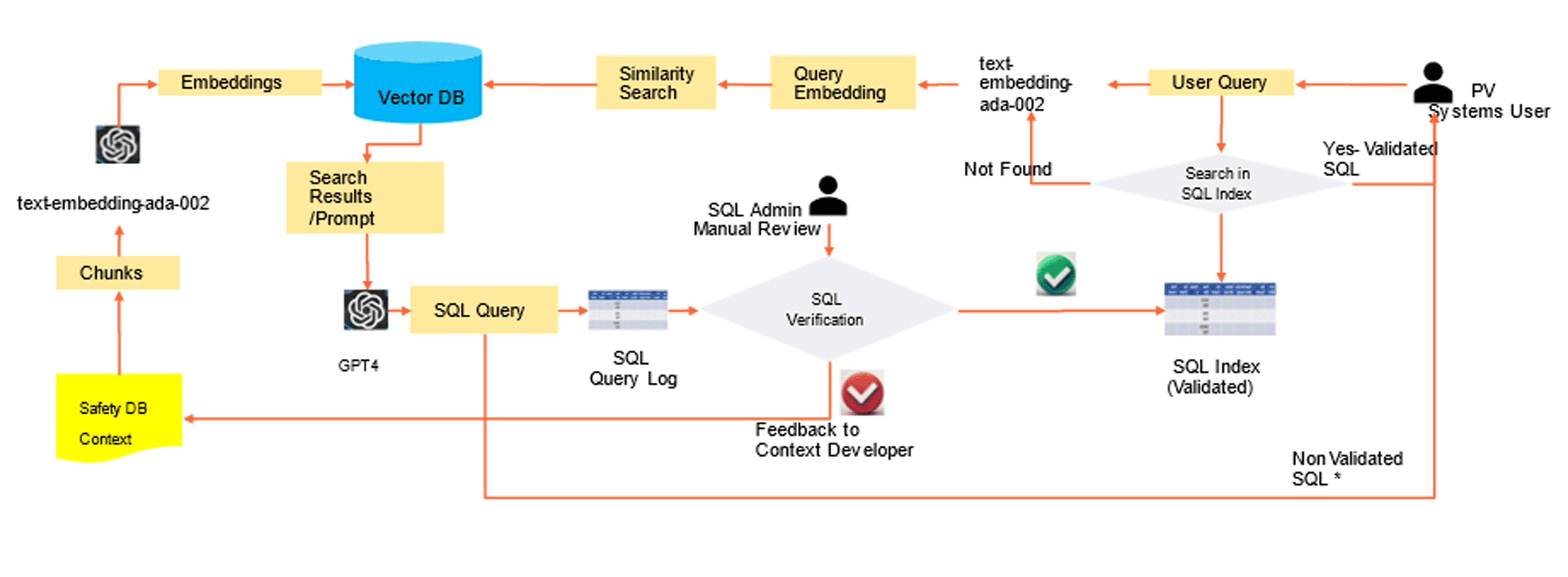}
				\caption{System Architectural Diagram}\label{fig_04}
			\end{figure}
			
			The business context document plays a central role in our system, providing vector embeddings that enhance query accuracy. Figure \ref{fig_04} illustrates how user queries are processed, embedded, and matched with relevant data, generating SQL queries based on both user input and contextual data from the document.

		\section{Results}
			
\begin{figure}[h]
				\centering
				\includegraphics[height=3.0in]{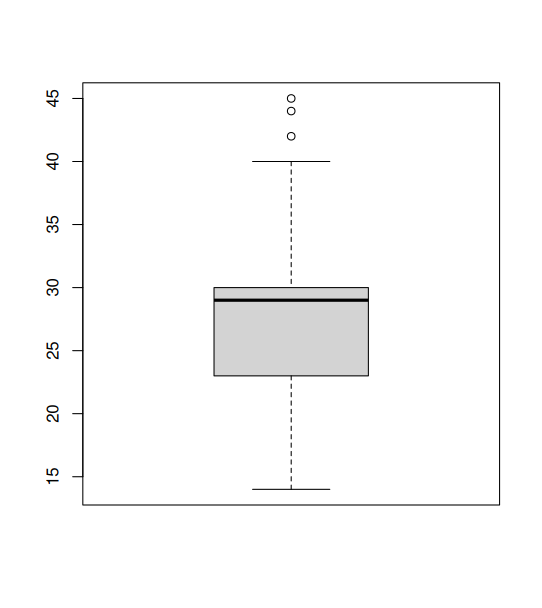}
				\caption{Boxplot of SQL Complexity Scores}\label{fig_05}
			\end{figure}

			Our experiment evaluated the impact of integrating a business context document with LLMs on the complexity and accuracy of SQL queries generated from NLQs. After applying our scoring algorithm to the generated SQL queries, we conducted a distribution analysis of the complexity scores. Figure \ref{fig_05} illustrates this distribution, categorizing SQL query complexity as ``low,'' ``medium,'' or ``high.'' These categories were based on scoring percentiles: queries scoring below the 25th percentile were classified as ``low,'' those within the interquartile range (IQR) as ``medium,'' and those above the 75th percentile as ``high.'' This method provides an objective measure of SQL query complexity, incorporating both subjective expertise and quantifiable metrics like table count.
			
			Of the 60 NLQs analyzed, 17 were categorized as ``low'' complexity, 31 as ``medium,'' and 12 as ``high.'' As outlined in the methods, the LLM-generated SQL queries were classified as `pass,' `partial success,' or `fail,' based on their accuracy. In Phase 1, where only the database schema was provided, the LLM achieved a pass rate of 8.3\%, with 78.3\% failing to generate valid SQL queries (Table \ref{dbresults}). This highlighted the challenges of query generation without additional context.
			
			\begin{table}[h]
				\caption{LLM Performance with DB Schema Only}\label{dbresults}%
				\begin{tabular}{@{}llllll@{}}
					\toprule
					& \multicolumn{3}{c}{\textbf{NLQ Complexity}} & & \\
					\textbf{Result} & Low  & Medium & High & Total & Percent \\
					\midrule
					Pass           & 3  & 2  & 0  & 5  & (8.3\%) \\
					Fail           & 11 & 27 & 9  & 47 & (78.3\%) \\
					Partial Pass   & 3  & 2  & 3  & 8  & (13.3\%) \\
					\midrule
					\textbf{Total} & 17 & 31 & 12 & 60 & \\
					\botrule
				\end{tabular}
			\end{table}
			
			The integration of the business context document in Phase 2 significantly improved performance, increasing the pass rate to 78.3\% (Table \ref{contxtresults}). This demonstrates the critical role of contextual knowledge in enhancing the LLM's ability to generate accurate SQL queries, particularly for complex NLQs. Additionally, when excluding high-complexity queries, the LLM achieved an 85.4\% pass rate, with 41 out of 48 low and medium complexity queries passing, and only 5 (10.4\%) resulting in failure.
			
			\begin{table}[h]
				\caption{LLM Performance with Business Context Document}\label{contxtresults}%
				\begin{tabular}{@{}llllll@{}}
					\toprule
					& \multicolumn{3}{c}{\textbf{NLQ Complexity}} & & \\
					\textbf{Result} & Low  & Medium & High & Total & Percent \\
					\midrule
					Pass           & 16 & 25 & 6  & 47 & (78.3\%) \\
					Fail           & 0  & 5  & 0  & 5  & (8.3\%) \\
					Partial Pass   & 1  & 1  & 6  & 8  & (13.3\%) \\
					\midrule
					\textbf{Total} & 17 & 31 & 12 & 60 & \\
					\botrule
				\end{tabular}
			\end{table}
			
			Statistical analysis using Fisher's Exact Test revealed a substantial improvement, with the p-value dropping from 0.1655 without the business context to 0.0006 with it, confirming the statistically significant performance boost provided by contextual information.
			
			The improvements were consistent across all complexity levels, highlighting the effectiveness of contextual priming in improving LLM SQL generation. Despite some remaining challenges with high-complexity queries, these results suggest that business context documents are a valuable tool for improving database interactions and LLM applications in data-intensive fields.
			
			In \textbf{Phase 3}, the schema was narrowed to essential tables without including the business context document. This experiment aimed to reduce ambiguity and assess baseline performance with a smaller set of relevant tables. The results showed modest improvements, reducing the failure rate from 78\% to 50\%, but many queries still required additional adjustments (Table \ref{narrowresults}).
			
			\begin{table}[h]
				\caption{Phase 3 LLM Performance with Narrowed Schema Definition}\label{narrowresults}%
				\begin{tabular}{@{}llllll@{}}
					\toprule
					& \multicolumn{3}{c}{\textbf{NLQ Complexity}} & & \\
					\textbf{Result} & Low  & Medium & High & Total & Percent \\
					\midrule
					Pass           & 4  & 2  & 0   & 6  & (10\%) \\
					Fail           & 6  & 18  & 6  & 30 & (50\%) \\
					Partial Pass   & 7  & 11  & 6  & 24 & (40\%) \\
					\midrule
					\textbf{Total} & 17 & 31 & 12 & 60 & \\
					\botrule
				\end{tabular}
			\end{table}
			
			Fisher’s Exact Test for Phase 3 yielded a p-value of 0.2373, indicating that narrowing the schema led to modest improvements but did not match the success achieved with the business context document. These findings emphasize that while schema optimization can help, the comprehensive insights provided by the business context document play a more critical role in enhancing the LLM's SQL query generation accuracy.
		
	\section{Discussion}
		
		Through our multi-phase analysis, this study has shown the significant impact of context-enriched LLMs in enhancing data retrieval from NLQs within PV and other data-intensive domains. By integrating OpenAI’s GPT-4 model with a business context document, we markedly improved the model's ability to generate syntactically precise and contextually relevant queries. This approach offers a promising pathway toward democratizing access to complex databases and enhancing the intuitiveness and efficiency of query formulation.
		
		Our findings indicate that augmenting the LLM with contextual knowledge substantially improves query generation accuracy. Specifically, the introduction of a business context document resulted in a success rate exceeding 78\% across a wide range of query complexities, highlighting the critical role of context in bridging the gap between natural language and database queries.
		
		Recent literature has identified similar challenges in text-to-SQL generation. For example, Qu et al. identify common issues such as schema-based and logic-based hallucinations \cite{qu2024before}. Their Task Alignment (TA) strategy aims to mitigate these hallucinations by aligning tasks with familiar contexts. Our approach, which incorporates a business context document, appears to effectively eliminate many of these hallucinations by providing detailed, domain-specific knowledge that guides the LLM in generating accurate SQL queries. This alignment reduces the risk of generating erroneous or irrelevant queries.

		Moreover, the CHESS framework's benchmarking using the BIRD database presents an incongruent point of comparison. BIRD-SQL\footnote{\url{https://bird-bench.github.io/}} contains hundreds of sub-datasets; however, it does not clearly indicate the complexity of these sub-datasets, which may not reflect the challenges posed by enterprise databases like ours. Unlike the BIRD benchmarks, our complexity algorithm explicitly documents how the level of complexity was determined. Our enterprise database, which again contains over 500 tables with an average of more than 50 columns per table, encompasses multiple ambiguities and complex relationships. The database represents over 5 million safety cases, with the largest table containing nearly 1.3 billion rows, and a single safety case may have as many as 100 products and 100 events reported. These characteristics significantly impact the conversion of NLQs to valid SQL queries for retrieving relevant safety data. This suggests that the complexity of real-world applications is not adequately represented in BIRD-SQL, highlighting the need for more representative benchmarking datasets.
		
		While our research seemingly marks a significant advancement in making informatics retrieval more accessible, enabling non-technical users to harness data-driven insights for more inclusive and efficient decision-making, it is essential to interpret our findings with caution. The efficacy of the business context document was assessed with a relatively small set of NLQs within a single enterprise database. The scalability and generalizability of our findings to other databases remain to be validated. Future research should aim to validate these results across broader datasets and diverse database architectures to fully understand the potential and limitations of our methodology.
		
		Further advancements are necessary. Our system identifies key areas for additional research and development, particularly in handling complex queries and resolving ambiguous user intents. Concerns related to scalability and implementation within large, dynamic enterprise environments highlight the need for future investigations to enhance the robustness and applicability of our methodology.
			
	\section{Conclusion}
		
		This study demonstrates the potential of leveraging Large Language Models (LLMs), specifically OpenAI's GPT-4, within a retrieval-augmented generation (RAG) framework to improve data retrieval from complex pharmacovigilance (PV) databases. Integrating a business context document significantly enhanced the model's ability to generate accurate and contextually relevant SQL queries from natural language queries (NLQs), with the success rate increasing from 8.3\% using the database schema alone to 78.3\% with the context document.
		
		Our findings emphasize the critical importance of contextual knowledge in bridging the gap between natural language and database queries, making data retrieval more intuitive and accessible for non-technical users. While these results are promising, further validation is needed to assess the scalability and generalizability of our approach across different databases and larger datasets.
		
		Future research should aim to validate these findings across a variety of database architectures and more extensive datasets, exploring the methodology's broader potential and limitations. Additionally, improving the system’s ability to handle complex queries and ambiguous user intents will be key to future development.
		
		In summary, this study introduces a novel approach for employing LLMs in data retrieval through NLQs, improving the accessibility and efficiency of PV data analysis. By integrating LLMs with a business context document, we propose a flexible pipeline that can be applied across multiple sectors, making complex databases more accessible to a wider range of users and supporting data-driven decision-making in various domains.

	\section{Acknowledgments}
		This research was supported in part by additional team members who helped with testing and development of the architectural framework to support context based LLMs, including Marcin Karwowski (GSK, Poland), Christie Roshan (GSK, UK) and Richard Barlow (GSK, USA), all members of the PV Systems team in Global Safety. 

	\section{Declarations}	
		\textbf{Data availability} The datasets generated during and/or analyzed during
			the current study are available from the corresponding author upon
			reasonable request. \\
		
		\textbf{Competing interests} The authors have no competing interests to
		declare that are relevant to the content of this article. \\
		
		\textbf{Conflict of interests} This manuscript has not been submitted to, nor is
		under review at, another journal or other publishing venue. \\
		
		\textbf{Funding} GlaxoSmithKline Biologicals SA covered all costs 
		associated with the conduct of the study and the development of the manuscript and the decision to publish the manuscript. JP, RK and AB are employees, and VC is a contractor, of the GSK group of companies. JP, RK and AB hold shares and/or options in the GSK group of companies.
	
	\bibliography{genai-sql-2024}

%% BioMed_Central_Bib_Style_v1.01

\begin{thebibliography}{21}
% BibTex style file: bmc-mathphys.bst (version 2.1), 2014-07-24
\ifx \bisbn   \undefined \def \bisbn  #1{ISBN #1}\fi
\ifx \binits  \undefined \def \binits#1{#1}\fi
\ifx \bauthor  \undefined \def \bauthor#1{#1}\fi
\ifx \batitle  \undefined \def \batitle#1{#1}\fi
\ifx \bjtitle  \undefined \def \bjtitle#1{#1}\fi
\ifx \bvolume  \undefined \def \bvolume#1{\textbf{#1}}\fi
\ifx \byear  \undefined \def \byear#1{#1}\fi
\ifx \bissue  \undefined \def \bissue#1{#1}\fi
\ifx \bfpage  \undefined \def \bfpage#1{#1}\fi
\ifx \blpage  \undefined \def \blpage #1{#1}\fi
\ifx \burl  \undefined \def \burl#1{\textsf{#1}}\fi
\ifx \doiurl  \undefined \def \doiurl#1{\url{https://doi.org/#1}}\fi
\ifx \betal  \undefined \def \betal{\textit{et al.}}\fi
\ifx \binstitute  \undefined \def \binstitute#1{#1}\fi
\ifx \binstitutionaled  \undefined \def \binstitutionaled#1{#1}\fi
\ifx \bctitle  \undefined \def \bctitle#1{#1}\fi
\ifx \beditor  \undefined \def \beditor#1{#1}\fi
\ifx \bpublisher  \undefined \def \bpublisher#1{#1}\fi
\ifx \bbtitle  \undefined \def \bbtitle#1{#1}\fi
\ifx \bedition  \undefined \def \bedition#1{#1}\fi
\ifx \bseriesno  \undefined \def \bseriesno#1{#1}\fi
\ifx \blocation  \undefined \def \blocation#1{#1}\fi
\ifx \bsertitle  \undefined \def \bsertitle#1{#1}\fi
\ifx \bsnm \undefined \def \bsnm#1{#1}\fi
\ifx \bsuffix \undefined \def \bsuffix#1{#1}\fi
\ifx \bparticle \undefined \def \bparticle#1{#1}\fi
\ifx \barticle \undefined \def \barticle#1{#1}\fi
\bibcommenthead
\ifx \bconfdate \undefined \def \bconfdate #1{#1}\fi
\ifx \botherref \undefined \def \botherref #1{#1}\fi
\ifx \url \undefined \def \url#1{\textsf{#1}}\fi
\ifx \bchapter \undefined \def \bchapter#1{#1}\fi
\ifx \bbook \undefined \def \bbook#1{#1}\fi
\ifx \bcomment \undefined \def \bcomment#1{#1}\fi
\ifx \oauthor \undefined \def \oauthor#1{#1}\fi
\ifx \citeauthoryear \undefined \def \citeauthoryear#1{#1}\fi
\ifx \endbibitem  \undefined \def \endbibitem {}\fi
\ifx \bconflocation  \undefined \def \bconflocation#1{#1}\fi
\ifx \arxivurl  \undefined \def \arxivurl#1{\textsf{#1}}\fi
\csname PreBibitemsHook\endcsname

%%% 1
\bibitem[\protect\citeauthoryear{Beninger}{2018}]{beninger2018pharmacovigilance}
\begin{barticle}
\bauthor{\bsnm{Beninger}, \binits{P.}}:
\batitle{Pharmacovigilance: an overview}.
\bjtitle{Clinical therapeutics}
\bvolume{40}(\bissue{12}),
\bfpage{1991}--\blpage{2004}
(\byear{2018})
\end{barticle}
\endbibitem

%%% 2
\bibitem[\protect\citeauthoryear{Brass and Goldberg}{2006}]{brass2006semantic}
\begin{barticle}
\bauthor{\bsnm{Brass}, \binits{S.}},
\bauthor{\bsnm{Goldberg}, \binits{C.}}:
\batitle{{Semantic errors in SQL queries: A quite complete list}}.
\bjtitle{Journal of Systems and Software}
\bvolume{79}(\bissue{5}),
\bfpage{630}--\blpage{644}
(\byear{2006})
\end{barticle}
\endbibitem

%%% 3
\bibitem[\protect\citeauthoryear{Sivarajkumar
  et~al.}{2024}]{sivarajkumar2024clinical}
\begin{botherref}
\oauthor{\bsnm{Sivarajkumar}, \binits{S.}},
\oauthor{\bsnm{Mohammad}, \binits{H.A.}},
\oauthor{\bsnm{Oniani}, \binits{D.}},
\oauthor{\bsnm{Roberts}, \binits{K.}},
\oauthor{\bsnm{Hersh}, \binits{W.}},
\oauthor{\bsnm{Liu}, \binits{H.}},
\oauthor{\bsnm{He}, \binits{D.}},
\oauthor{\bsnm{Visweswaran}, \binits{S.}},
\oauthor{\bsnm{Wang}, \binits{Y.}}:
Clinical information retrieval: A literature review.
Journal of Healthcare Informatics Research,
1--40
(2024)
\end{botherref}
\endbibitem

%%% 4
\bibitem[\protect\citeauthoryear{Mao et~al.}{2023}]{mao2023two}
\begin{barticle}
\bauthor{\bsnm{Mao}, \binits{Y.}},
\bauthor{\bsnm{Miller}, \binits{R.A.}},
\bauthor{\bsnm{Bodenreider}, \binits{O.}},
\bauthor{\bsnm{Nguyen}, \binits{V.}},
\bauthor{\bsnm{Fung}, \binits{K.W.}}:
\batitle{{Two complementary AI approaches for predicting UMLS semantic group
  assignment: heuristic reasoning and deep learning}}.
\bjtitle{Journal of the American Medical Informatics Association}
\bvolume{30}(\bissue{12}),
\bfpage{1887}--\blpage{1894}
(\byear{2023})
\end{barticle}
\endbibitem

%%% 5
\bibitem[\protect\citeauthoryear{Talaei et~al.}{2024}]{talaei2024chess}
\begin{botherref}
\oauthor{\bsnm{Talaei}, \binits{S.}},
\oauthor{\bsnm{Pourreza}, \binits{M.}},
\oauthor{\bsnm{Chang}, \binits{Y.-C.}},
\oauthor{\bsnm{Mirhoseini}, \binits{A.}},
\oauthor{\bsnm{Saberi}, \binits{A.}}:
{CHESS: Contextual Harnessing for Efficient SQL Synthesis}
(2024)
\end{botherref}
\endbibitem

%%% 6
\bibitem[\protect\citeauthoryear{Taipalus et~al.}{2018}]{taipalus2018errors}
\begin{barticle}
\bauthor{\bsnm{Taipalus}, \binits{T.}},
\bauthor{\bsnm{Siponen}, \binits{M.}},
\bauthor{\bsnm{Vartiainen}, \binits{T.}}:
\batitle{{Errors and complications in SQL query formulation}}.
\bjtitle{ACM Transactions on Computing Education (TOCE)}
\bvolume{18}(\bissue{3}),
\bfpage{1}--\blpage{29}
(\byear{2018})
\end{barticle}
\endbibitem

%%% 7
\bibitem[\protect\citeauthoryear{Zloof}{1977}]{zloof1977query}
\begin{barticle}
\bauthor{\bsnm{Zloof}, \binits{M.M.}}:
\batitle{{Query-by-example: A data base language}}.
\bjtitle{IBM systems Journal}
\bvolume{16}(\bissue{4}),
\bfpage{324}--\blpage{343}
(\byear{1977})
\end{barticle}
\endbibitem

%%% 8
\bibitem[\protect\citeauthoryear{Ramakrishnan and
  Gehrke}{2002}]{ramakrishnan2002database}
\begin{bbook}
\bauthor{\bsnm{Ramakrishnan}, \binits{R.}},
\bauthor{\bsnm{Gehrke}, \binits{J.}}:
\bbtitle{Database Management Systems}.
\bpublisher{McGraw-Hill, Inc.},
\blocation{New York, NY}
(\byear{2002})
\end{bbook}
\endbibitem

%%% 9
\bibitem[\protect\citeauthoryear{Dauphin et~al.}{2017}]{dauphin2017language}
\begin{bchapter}
\bauthor{\bsnm{Dauphin}, \binits{Y.N.}},
\bauthor{\bsnm{Fan}, \binits{A.}},
\bauthor{\bsnm{Auli}, \binits{M.}},
\bauthor{\bsnm{Grangier}, \binits{D.}}:
\bctitle{Language modeling with gated convolutional networks}.
In: \bbtitle{International Conference on Machine Learning},
pp. \bfpage{933}--\blpage{941}
(\byear{2017}).
\bcomment{PMLR}
\end{bchapter}
\endbibitem

%%% 10
\bibitem[\protect\citeauthoryear{Li et~al.}{2023}]{li2023can}
\begin{botherref}
\oauthor{\bsnm{Li}, \binits{J.}},
\oauthor{\bsnm{Hui}, \binits{B.}},
\oauthor{\bsnm{Qu}, \binits{G.}},
\oauthor{\bsnm{Li}, \binits{B.}},
\oauthor{\bsnm{Yang}, \binits{J.}},
\oauthor{\bsnm{Li}, \binits{B.}},
\oauthor{\bsnm{Wang}, \binits{B.}},
\oauthor{\bsnm{Qin}, \binits{B.}},
\oauthor{\bsnm{Cao}, \binits{R.}},
\oauthor{\bsnm{Geng}, \binits{R.}}, et al.:
{Can LLM Already Serve as A Database Interface? A Big Bench for Large-Scale
  Database Grounded Text-to-SQLs}.
arXiv preprint arXiv:2305.03111
(2023)
\end{botherref}
\endbibitem

%%% 11
\bibitem[\protect\citeauthoryear{Gao et~al.}{2023}]{gao2023text}
\begin{botherref}
\oauthor{\bsnm{Gao}, \binits{D.}},
\oauthor{\bsnm{Wang}, \binits{H.}},
\oauthor{\bsnm{Li}, \binits{Y.}},
\oauthor{\bsnm{Sun}, \binits{X.}},
\oauthor{\bsnm{Qian}, \binits{Y.}},
\oauthor{\bsnm{Ding}, \binits{B.}},
\oauthor{\bsnm{Zhou}, \binits{J.}}:
{Text-to-SQL Empowered by Large Language Models: A Benchmark Evaluation}.
arXiv preprint arXiv:2308.15363
(2023)
\end{botherref}
\endbibitem

%%% 12
\bibitem[\protect\citeauthoryear{Painter et~al.}{2023}]{painterLLM2023}
\begin{botherref}
\oauthor{\bsnm{Painter}, \binits{J.L.}},
\oauthor{\bsnm{Mahaux}, \binits{O.}},
\oauthor{\bsnm{Vanini}, \binits{M.}},
\oauthor{\bsnm{Kara}, \binits{V.}},
\oauthor{\bsnm{Roshan}, \binits{C.}},
\oauthor{\bsnm{Karwowski}, \binits{M.}},
\oauthor{\bsnm{Chalamalasetti}, \binits{V.R.}},
\oauthor{\bsnm{Bate}, \binits{A.}}:
{Enhancing Drug Safety Documentation Search Capabilities with Large Language
  Models: A User-Centric Approach}.
2023 International Conference on Computational Science and Computational
  Intelligence (CSCI)
(2023).
CSCI
\end{botherref}
\endbibitem

%%% 13
\bibitem[\protect\citeauthoryear{Roberson et~al.}{2024}]{roberson2024analyzing}
\begin{botherref}
\oauthor{\bsnm{Roberson}, \binits{R.}},
\oauthor{\bsnm{Kaki}, \binits{G.}},
\oauthor{\bsnm{Trivedi}, \binits{A.}}:
{Analyzing the Effectiveness of Large Language Models on Text-to-SQL
  Synthesis}.
arXiv preprint arXiv:2401.12379
(2024)
\end{botherref}
\endbibitem

%%% 14
\bibitem[\protect\citeauthoryear{Zhang et~al.}{2024}]{zhang2024structure}
\begin{botherref}
\oauthor{\bsnm{Zhang}, \binits{Q.}},
\oauthor{\bsnm{Dong}, \binits{J.}},
\oauthor{\bsnm{Chen}, \binits{H.}},
\oauthor{\bsnm{Li}, \binits{W.}},
\oauthor{\bsnm{Huang}, \binits{F.}},
\oauthor{\bsnm{Huang}, \binits{X.}}:
{Structure Guided Large Language Model for SQL Generation}.
arXiv preprint arXiv:2402.13284
(2024)
\end{botherref}
\endbibitem

%%% 15
\bibitem[\protect\citeauthoryear{Qu et~al.}{2024}]{qu2024before}
\begin{botherref}
\oauthor{\bsnm{Qu}, \binits{G.}},
\oauthor{\bsnm{Li}, \binits{J.}},
\oauthor{\bsnm{Li}, \binits{B.}},
\oauthor{\bsnm{Qin}, \binits{B.}},
\oauthor{\bsnm{Huo}, \binits{N.}},
\oauthor{\bsnm{Ma}, \binits{C.}},
\oauthor{\bsnm{Cheng}, \binits{R.}}:
{Before Generation, Align it! A Novel and Effective Strategy for Mitigating
  Hallucinations in Text-to-SQL Generation}.
arXiv preprint arXiv:2405.15307
(2024)
\end{botherref}
\endbibitem

%%% 16
\bibitem[\protect\citeauthoryear{Simon and Pataki}{2010}]{simon2010sql}
\begin{bchapter}
\bauthor{\bsnm{Simon}, \binits{M.}},
\bauthor{\bsnm{Pataki}, \binits{N.}}:
\bctitle{{SQL Code Complexity Analysis}}.
In: \bbtitle{Proceedings of the 8th International Conference of Applied
  Informatics}
(\byear{2010})
\end{bchapter}
\endbibitem

%%% 17
\bibitem[\protect\citeauthoryear{Brown et~al.}{2020}]{brown2020language}
\begin{barticle}
\bauthor{\bsnm{Brown}, \binits{T.}},
\bauthor{\bsnm{Mann}, \binits{B.}},
\bauthor{\bsnm{Ryder}, \binits{N.}},
\bauthor{\bsnm{Subbiah}, \binits{M.}},
\bauthor{\bsnm{Kaplan}, \binits{J.D.}},
\bauthor{\bsnm{Dhariwal}, \binits{P.}},
\bauthor{\bsnm{Neelakantan}, \binits{A.}},
\bauthor{\bsnm{Shyam}, \binits{P.}},
\bauthor{\bsnm{Sastry}, \binits{G.}},
\bauthor{\bsnm{Askell}, \binits{A.}}, \betal:
\batitle{Language models are few-shot learners}.
\bjtitle{Advances in neural information processing systems}
\bvolume{33},
\bfpage{1877}--\blpage{1901}
(\byear{2020})
\end{barticle}
\endbibitem

%%% 18
\bibitem[\protect\citeauthoryear{Xie et~al.}{2021}]{xie2021explanation}
\begin{botherref}
\oauthor{\bsnm{Xie}, \binits{S.M.}},
\oauthor{\bsnm{Raghunathan}, \binits{A.}},
\oauthor{\bsnm{Liang}, \binits{P.}},
\oauthor{\bsnm{Ma}, \binits{T.}}:
An explanation of in-context learning as implicit bayesian inference.
arXiv preprint arXiv:2111.02080
(2021)
\end{botherref}
\endbibitem

%%% 19
\bibitem[\protect\citeauthoryear{Radford et~al.}{2019}]{radford2019language}
\begin{barticle}
\bauthor{\bsnm{Radford}, \binits{A.}},
\bauthor{\bsnm{Wu}, \binits{J.}},
\bauthor{\bsnm{Child}, \binits{R.}},
\bauthor{\bsnm{Luan}, \binits{D.}},
\bauthor{\bsnm{Amodei}, \binits{D.}},
\bauthor{\bsnm{Sutskever}, \binits{I.}}, \betal:
\batitle{Language models are unsupervised multitask learners}.
\bjtitle{OpenAI blog}
\bvolume{1}(\bissue{8}),
\bfpage{9}
(\byear{2019})
\end{barticle}
\endbibitem

%%% 20
\bibitem[\protect\citeauthoryear{Lewis et~al.}{2020}]{lewis2020retrieval}
\begin{barticle}
\bauthor{\bsnm{Lewis}, \binits{P.}},
\bauthor{\bsnm{Perez}, \binits{E.}},
\bauthor{\bsnm{Piktus}, \binits{A.}},
\bauthor{\bsnm{Petroni}, \binits{F.}},
\bauthor{\bsnm{Karpukhin}, \binits{V.}},
\bauthor{\bsnm{Goyal}, \binits{N.}},
\bauthor{\bsnm{K{\"u}ttler}, \binits{H.}},
\bauthor{\bsnm{Lewis}, \binits{M.}},
\bauthor{\bsnm{Yih}, \binits{W.-t.}},
\bauthor{\bsnm{Rockt{\"a}schel}, \binits{T.}}, \betal:
\batitle{Retrieval-augmented generation for knowledge-intensive {NLP} tasks}.
\bjtitle{Advances in Neural Information Processing Systems}
\bvolume{33},
\bfpage{9459}--\blpage{9474}
(\byear{2020})
\end{barticle}
\endbibitem

%%% 21
\bibitem[\protect\citeauthoryear{Topsakal and
  Akinci}{2023}]{topsakal2023creating}
\begin{bchapter}
\bauthor{\bsnm{Topsakal}, \binits{O.}},
\bauthor{\bsnm{Akinci}, \binits{T.C.}}:
\bctitle{{Creating large language model applications utilizing langchain: A
  primer on developing LLM apps fast}}.
In: \bbtitle{International Conference on Applied Engineering and Natural
  Sciences},
vol. \bseriesno{1},
pp. \bfpage{1050}--\blpage{1056}
(\byear{2023})
\end{bchapter}
\endbibitem

\end{thebibliography}
	
\end{document}